\documentclass{article}
\usepackage{ijcai21}

\usepackage{times}
\usepackage{soul}
\usepackage{url}
\usepackage[hidelinks]{hyperref}
\usepackage[utf8]{inputenc}
\usepackage[small]{caption}
\usepackage{graphicx}
\usepackage{amsmath}
\usepackage{amsthm}
\usepackage{booktabs}
\usepackage{algorithm}
\usepackage{algorithmic}
\usepackage{CJKutf8}
\usepackage{makecell}
\usepackage{multirow}
\usepackage{graphicx}
\usepackage{amsmath}
\usepackage{amssymb}
\usepackage{xcolor}
\newcolumntype{L}[1]{>{\raggedright\let\newline\\\arraybackslash\hspace{0pt}}m{#1}}
\usepackage{bm}

\newcommand*{\affaddr}[1]{#1} 
\newcommand*{\affmark}[1][*]{\textsuperscript{#1}}
\newcommand*{\email}[1]{\texttt{#1}}

\title{Enhancing Dialogue Generation via Multi-Level Contrastive Learning\thanks{This work was done when Xin Li was an intern as Tencent AI Lab.}}

\author{

Xin Li\affmark[1], Piji Li\affmark[2], Yan Wang\affmark[2], Xiaojiang Liu\affmark[2] and Wai Lam\affmark[1] \\
\affaddr{\affmark[1]The Chinese University of Hong Kong}\\
\affaddr{\affmark[2]Tencent AI Lab}\\
\email{\{lixin,wlam\}@se.cuhk.edu.hk}\\
\email{\{pijili,brandenwang,kieranliu\}@tencent.com}\\
}

\begin{document}

\maketitle

\begin{abstract}
Most of the existing works for dialogue generation are data-driven models trained directly on corpora crawled from websites. They mainly focus on improving the model architecture to produce better responses but pay little attention to considering the quality of the training data contrastively. In this paper, we propose a multi-level contrastive learning paradigm to model the fine-grained quality of the responses with respect to the query. A Rank-aware Calibration (RC) network is designed to construct the multi-level contrastive optimization objectives. Instead of emphasising the sentence-level quality, which may erroneously encourage/suppress the generation of uninformative/informative words, we design an exquisite token-level strategy for estimating the instance loss more accurately. On the other hand, we build a Knowledge Inference (KI) component to capture the keyword knowledge from the reference during training and exploit such information to encourage the generation of informative words. We evaluate the proposed model on a carefully annotated dialogue dataset and the results suggest that our model can generate more relevant and diverse responses compared to the baseline models.
\end{abstract}

\section{Introduction}
Response generation for dialogue systems has stimulated great interests for researchers recently~\cite{ritter-etal-2011-data,shang-etal-2015-neural,vinyals2015neural}. The core idea of dialogue generation is to formulate the task as a sequence translation problem and translate the query to a response. One common neural model is the sequence-to-sequence (S2S) encoder-decoder framework~\cite{cho-etal-2014-learning,sutskever2014sequence,bahdanau2015neural}. Many approaches have been proposed to improve the basic S2S model for better human-computer conversation performance~\cite{li-etal-2016-diversity,li-etal-2017-adversarial,xing2017topic,shao-etal-2017-generating,xu-etal-2017-neural,zhou2017mechanism,pei-li-2018-s2spmn,wang2018chat,gao2019generating,zhang-etal-2020-dialogpt,he-glass-2020-negative,tian-etal-2020-response}. 

Despite their popularity, these approaches assume that each training sample, namely, query-response pair, contributes equally to the model and ignore the consideration of different response quality contrastively. Table~\ref{tab:example} depicts some example responses for a particular query in a dialogue dataset. Both of the first and the second response are relevant to the query but the first one is obviously better when considering informativeness and interestingness. The third response is acceptable for the conversation but quite universal, meaning that, it can also be used to answer other queries. Thus, its quality is not as good as that of the first two responses. The fourth response is poor since it directly copies part of the query. Although the fourth response is not acceptable, it is still better when compared with the fifth response, which is completely irrelevant to the query.

\begin{table}[!t]
    \centering
    \resizebox{1.0\columnwidth}{!}{
    \begin{tabular}{L{2.5cm}|L{5cm}|L{4cm}}
    \Xhline{3\arrayrulewidth}
       \textbf{Query} & \textbf{Response} & \textbf{Quality} \\ \hline
       & \begin{CJK}{UTF8}{gbsn}1. 下雨虽然带来很多不便, 可太阳也好无情啊, 就怕晒\end{CJK} & \multirow{2}{*}{Relevant and interesting} \\ 
       & Although rains are inconvenient, sunny day is no better than rainy day since I am afraid of the sun. &  \\ \cline{2-3}
       & \begin{CJK}{UTF8}{gbsn}2. 最讨厌下雨天!\end{CJK} & \multirow{2}{*}{Relevant but simple}. \\ 
       & I hate rainy days! & \\ \cline{2-3}
       \begin{CJK}{UTF8}{gbsn}又开始下雨了\end{CJK} & \begin{CJK}{UTF8}{gbsn}3. 你现在青岛还是广州？\end{CJK} & \multirow{2}{*}{Acceptable but universal} \\ 
       It started to rain again & Are you in Qingdao or Guangzhou now? &  \\ \cline{2-3}
       & \begin{CJK}{UTF8}{gbsn}4. 嗯, 又下雨了\end{CJK} & \multirow{2}{*}{Quiet boring} \\ 
       & Yes, it rains again. & \\ \cline{2-3}
       & \begin{CJK}{UTF8}{gbsn}5. 这所大学和烈士陵园很近 \end{CJK} & \multirow{2}{*}{Irrelevant} \\ 
       & The university is near martyrs cemetery. & \\ 
    \Xhline{3\arrayrulewidth}
    \end{tabular}}
    \caption{Example of a sample from training-set: query with responses in different quality.}
    \label{tab:example}
    \vspace{-5mm}
\end{table}

Some initial attempts~\cite{shang2018learning,liu-etal-2018-towards-less} have been conducted to consider the quality of the training data. Following the idea of instance weighting~\cite{jiang-zhai-2007-instance}, \cite{shang2018learning} pre-train a calibration network to calculate the response quality score for each training sample (i.e., query-response pair) and update the model with the weighted combination of the sample loss. Similarly, \cite{liu-etal-2018-towards-less} estimate the instance score based on the corpus-level n-gram co-occurrence and the length of the response. Both of them are simple to implement but they still have some limitations: (1) The calibration network is only trained on relevant responses and irrelevant responses from other queries and therefore cannot capture the fine-grained response quality, as exemplified in Table~\ref{tab:example}; (2) The instance weighting strategy treats all tokens in the response as equal importance to the query by assigning them with the same quality score, which may erroneously
encourage the generation of some uninformative words in the relevant responses (e.g., ``since'' and ``afraid'' in the first response in Table~\ref{tab:example}).

To tackle the issues mentioned above, we introduce the Contrastive Learning paradigm~\cite{hadsell2006dimensionality,he2019momentum,chen2020simple,iter2020pretraining} to model the multi-level fine-grained quality of the responses with respect to the query.
Specifically, we develop a Rank-aware Calibration (RC) network aiming for modeling the fine-grained quality and characterizing the response properties (e.g., relevance and informativeness) that will affect the conversation experience with a multi-scale response quality score. The rank-aware calibrator adopts the strategies of pointwise regression and pairwise ranking for gauging the quality of the query-response pair. Besides, to address the second limitation aforementioned, we design a more exquisite strategy to consider the different importance of tokens instead of simply scaling the training sample loss with the response-level quality score. Concretely, we propose to conditionally sample a response via Monte-Carlo Rollout~\cite{yu2017seqgan,lin2017adversarial} for each gold standard response token and deem the quality scores of the sampled responses as the importance of the tokens in the sample loss estimation.

It is also observed that some meaningful words such as ``university'' and ``martyrs cemetery'' in the fifth response in Table~\ref{tab:example} are very likely to receive low quality scores due to the irrelevance to the query. Thus, we propose \textbf{K}nowledge \textbf{I}nference (KI) component to explicitly encourage the generation of the informative tokens in the gold standard responses. This component firstly associates the query and the decoder hidden representation with the memories of the informative tokens and then incorporates the summarized memories into each decoding step.

In summary, our contributions are as follows:

\indent $\bullet$ To enhance the performance of dialogue generation, we propose a multi-level contrastive learning paradigm to model the fine-grained quality of the responses with respect to the query.\\
\indent $\bullet$ We propose a \textbf{R}ank-aware \textbf{C}alibration (RC) network to construct the multi-level contrastive objectives. We further design a strategy to calibrate the model training with token-level quality information.\\
\indent $\bullet$ We propose to reconsider the generation of some informative words erroneously punished by the calibrator via a tailor-made \textbf{K}nowledge \textbf{I}nference (KI) component. \\
\indent $\bullet$ We build a dataset with fine-grained response annotations and conduct extensive evaluations. The experimental results validate the effectiveness of the proposed framework. 

Code and the labelled dataset will be public to facilitate the research.

\begin{figure*}[!t]
    \centering
    \includegraphics[width=.7\textwidth]{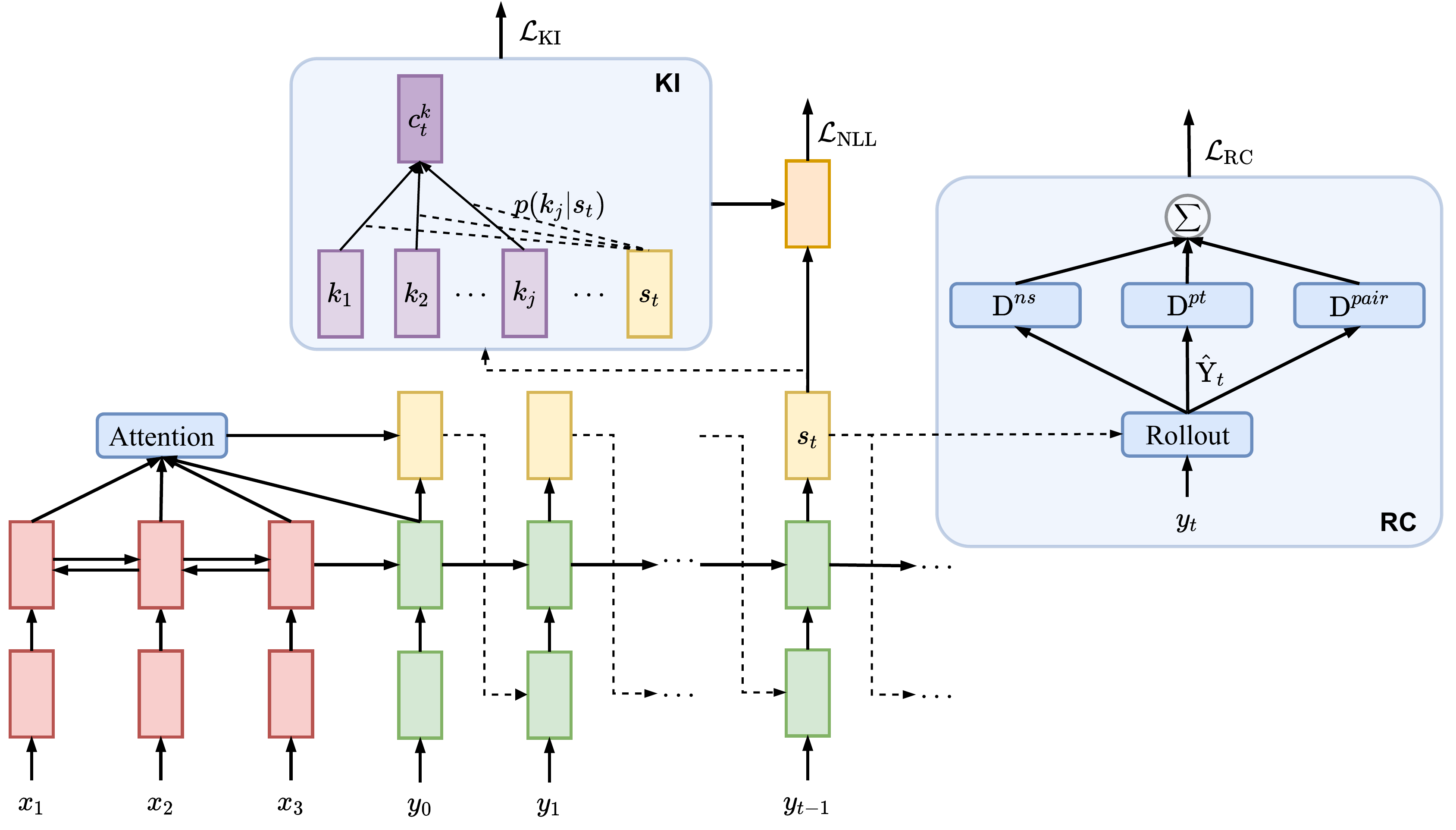}
    \caption{Architecture of the proposed model. Note that the components \textbf{KI} and \textbf{RC} are disabled during inference.}
    \label{fig:architecture}
    \vspace{-5mm}
\end{figure*}

\section{Model}
\subsection{Overview}
Given a query-response pair ($\mathrm{\bf X}$, $\mathrm{\bf Y}$), where the query word sequence $\mathrm{\bf X} = \{x_1, x_2, \cdots, x_n\}$ and the response word sequence $\mathrm{\bf Y}=\{y_1, y_2, \cdots, y_m\}$. The aim of the dialogue generation task is to maximize the conditional probability $p(\mathrm{\bf Y}|\mathrm{\bf X})$ and generate the relevant and meaningful responses with respect to the query.

As shown in Figure~\ref{fig:architecture}, the model proposed in this paper consist of three major components: the backbone S2S model, a rank-aware calibration network (RC) for response quality estimation, and a knowledge inference component (KI) for encouraging the generation of informative words.
For RC-based response quality estimation, we apply Monte-Carlo Rollout with the policy network $G^{\beta}$ to perform response sampling conditioned on the current token $y_t$ and feed the sampled response $\mathrm{\bf \hat{Y}}_t \in \mathbb{R}^m$ to the RC, where three contrastive quality estimation components $\mathrm{D}^{ns}$, $\mathrm{D}^{pt}$ and $\mathrm{D}^{pair}$, namely, negative sampling component, pointwise ranking component, and pairwise ranking component are included. The produced quality scores of the $\mathrm{\bf \hat{Y}}_t$ are $\mathrm{q}^{ns}$, $\mathrm{q}^{pt}$, $\mathrm{q}^{pair}$ respectively.
For knowledge inference (KI), a knowledge memory $\mathrm{\bf K} \in \mathbb{R}^{n_k \times \mathrm{dim}_k}$ and a knowledge attention layer are designed. $n_k$ refers to the number of the knowledge memories for the informative tokens.
The knowledge summary ${\bf c}^{k}_t$ will also serve as the input for the $t$-th decoding step.

\subsection{Response Generation}
The backbone generation model is an attention-based sequence-to-sequence framework. Specifically, in encoding, we employ a bi-directional LSTM~\cite{hochreiter1997long} to map the query input $x_i$ to the distributed hidden representation ${\bm h}_i \in \mathbb{R}^{2\mathrm{dim}_h}$ ($i \in [1, n]$) as follows:
\begin{equation}
\setlength{\abovedisplayskip}{3pt}
\setlength{\belowdisplayskip}{3pt}
    {\bm h}_i = [\overrightarrow{\text{LSTM}}(x_i,\overrightarrow{{\bm h}}_{i-1}); \overleftarrow{\text{LSTM}}(x_i,\overleftarrow{{\bm h}}_{i+1})]
\end{equation}
where the operation of an LSTM unit on $x_t$ is denoted as $\text{LSTM}(x_i,h_{i-1})$. In decoding, the computational formula of the decoding hidden representation ${\bf s}_t$ at the $t$-th time step is below:
\begin{equation}
\setlength{\abovedisplayskip}{3pt}
\setlength{\belowdisplayskip}{3pt}
    {\bm s}_t = f(y_{t-1}, {\bm s}_{t-1}, {\bm c}_t)
\end{equation}
The context vector ${\bm c}_t$ is computed as the weighted combination of the encoder hidden memories $\mathrm{\bf H} = \{{\bm h}_1,\cdots,{\bm h}_n\}$ along with the attention signal ${\bm a}_t \in \mathbb{R}^{n}$. Then ${\bm s}_t$ is fed to a softmax layer to calculate the probability distribution of the candidate words:
\begin{equation}
\setlength{\abovedisplayskip}{3pt}
\setlength{\belowdisplayskip}{3pt}
    p(\mathrm{\bf Y}|\mathrm{\bf X}) = \prod^{m}_{t=1} p(y_t|\mathrm{\bf X},y_{1:t-1})
    \label{eq:likelihood}
\end{equation}

\subsection{Rank-Aware Quality Estimation}
As shown in Figure~\ref{fig:architecture}, the \textbf{R}ank-aware \textbf{C}alibrator (RC) model is composed of three components: $\mathrm{D}^{ns}$, $\mathrm{D}^{pt}$ and $\mathrm{D}^{pair}$. Each of the component contains a scoring network for calculating the quality score, either performing semantic matching or learning a joint semantic representation for a query and a response. The matching-based scoring network $g_{\mathrm{M}}$ maps a query and a response to vector representations and employs a bi-linear layer to measure the semantic relatedness between these two representations. The joint-representation-based scoring network $g_{\mathrm{J}}$ learns the joint semantic representation for the query and the response and utilize a non-linear transformation layer to obtain the quality score. To exploit the fine-grained response quality, we propose three ranking strategies, namely, \textbf{Negative Sampling}, \textbf{Pointwise Ranking} and \textbf{Pairwise Ranking}, to pre-train the components $\mathrm{D}^{ns}$, $\mathrm{D}^{pt}$ and $\mathrm{D}^{pair}$ respectively.
\paragraph{\textbf{Negative Sampling}}Negative sampling is the strategy for optimizing the model $\mathrm{D}^{ns}$. Specifically, for each query response pair ($\mathrm{\bf X}$, $\mathrm{\bf Y}$), we sample a response $\mathrm{\bf Y}^{-}$ from other queries in the training dataset to construct a negative sample ($\mathrm{\bf X}$, $\mathrm{\bf Y}^{-}$). Then, we augment the original training dataset with the sampled negative data and train the scoring network by maximizing the following objective:
\begin{equation}
\small
    \mathit{L}^{ns} = \mathrm{b} * g^{ns}_{\mathrm{J}}(\mathrm{\bf X}, \mathrm{\bf Y}^{*}) + (1 - \mathrm{b}) * (1-g^{ns}_{\mathrm{J}}(\mathrm{\bf X}, \mathrm{\bf Y}^{*}))
\end{equation}
where $g^{ns}_{\mathrm{J}}$ denotes the score obtained based on the joint representation of $\mathrm{\bf X}$ and $\mathrm{\bf Y}^{*}$. The binary variable $\mathrm{b}$ is set as 1 if $\mathrm{\bf Y}^{*}$ is from the original training dataset. Otherwise, $\mathrm{b}$ is 0.  Optimizing this classification-based objective can rank the relevant responses higher than the irrelevant ones.
\paragraph{\textbf{Pointwise Ranking}} As mentioned in the Introduction section, only differentiating relevant responses from the irrelevant responses is not enough. Since the multi-level quality annotations are available in the training dataset, one direct way for modeling the fine-grained response quality is to minimize the gap between the predicted quality scores and the gold standard multi-level scores. Thus, we propose a regression-based criterion for optimizing the model $\mathrm{D}^{pt}$. The objective function is given by:
\begin{equation}
\setlength{\abovedisplayskip}{3pt}
\setlength{\belowdisplayskip}{3pt}
    \mathit{L}^{pt} = \left(\mathrm{q}^g - g^{pt}_{\mathrm{M}}(\mathrm{\bf X}, \mathrm{\bf Y})\right)^2
\end{equation}
where $\mathrm{q}^g$ is the gold-standard quality score for the query-response pair ($\mathrm{\bf X}, \mathrm{\bf Y}$) and $g^{pt}_{\mathrm{M}}$ is the matching-based scoring function. With this objective, the $\mathrm{D}^{pt}$ is enforced to pay attention to different levels of quality scores rather than just 0-1 relevance. In other words, $\mathrm{D}^{pt}$ can exploit how good the response $\mathrm{\bf Y}$ is apart from its relevance/irrelevance to the query $\mathrm{\bf X}$.
\paragraph{\textbf{Pairwise Ranking}} The pointwise ranking strategy adopted in $\mathrm{D}^{pt}$ is to directly model the absolute quality scores. Compared to the pointwise ranking, the pairwise ranking strategy implicitly exploits the multi-level quality scores by learning to assign preference for the pre-built paired responses $\mathrm{\bf Y}^{++}$ and $\mathrm{\bf Y}^{+}$ from the same query $\mathrm{\bf X}$. Here, $\mathrm{\bf Y}^{++}$ and $\mathrm{\bf Y}^{+}$ denote the responses belonging to the adjacent quality levels and $\mathrm{\bf Y}^{++}$ is the better one. To optimize the scoring component $g^{pair}_{\mathrm{M}}$ in $\mathrm{D}^{pair}$, the pairwise ranking strategy rewards the better response in the ranking pair and penalizes the other one so that the margin of their quality estimations is as close as possible to the pre-specified value $\Delta$. The loss function is as follows:
\begin{equation}
\setlength{\abovedisplayskip}{3pt}
\setlength{\belowdisplayskip}{3pt}
    \small
    \mathit{L}^{pair} = \max\left(0, \Delta-g^{pair}_{\mathrm{M}}(\mathrm{\bf X}, \mathrm{\bf Y}^{++})+g^{pair}_{\mathrm{M}}(\mathrm{\bf X}, \mathrm{\bf Y}^{+})\right)
\end{equation}
where the score function $g^{pair}_{\mathrm{M}}$ calculates the quality scores of the response via matching the semantic representations of the response and the query. Then, we traverse all of the possible response pairs under the query $\mathrm{\bf X}$ to make the model $\mathrm{D}^{pair}$ aware of different quality levels.

Intuitively, we can utilize the RC model to estimate the quality of each training sample and leverage the quality scores to adjust the weight of the training sample loss. However, as mentioned above, such response-level strategy ignores the consideration of the importance of different tokens in the same response. In this paper, we give a solution that performing token-level calibration with the fine-grained quality estimation.

\subsection{Token-Level Calibration}
The aim of the vanilla sequence-to-sequence (S2S) model is to minimize the following negative log-likelihood (NLL) objective function:
\begin{equation}
\setlength{\abovedisplayskip}{3pt}
\setlength{\belowdisplayskip}{3pt}
    \mathcal{L}_{\text{NLL}} =\frac{1}{|\mathbb{D}|} \sum_{(\mathrm{\bf X}, \mathrm{\bf Y}) \in \mathbb{D}} -\log(p(\mathrm{\bf Y}|\mathrm{\bf X}))
\end{equation}
where $\mathbb{D}$ denotes the training dataset or a training batch and $p(\mathrm{\bf Y}|\mathrm{\bf X})$ is the conditional generation probability given by Eq~\ref{eq:likelihood}.

Recall that response-level quality estimation treats all of the tokens in the response as equal importance and cannot accurately reflect the importance of the token in the response, which may erroneously encourage the generation of some uninformative words.
In order to integrate the quality information more accurately, we propose to calibrate the S2S training with token-level quality information. 
Specifically, we propose to measure the quality of the sampled response conditioned on each token and deem the quality score estimated by rank-aware calibrator as the importance of the corresponding token in the sample loss estimation.

Given the training sample $\mathrm{\bf X}, \mathrm{\bf Y}$ and the gold standard response token $y_t$ at the $t$-th decoding step, we firstly apply Monte-Carlo search with a rollout policy $G^{\beta}$ (see the ``Rollout'' component in Figure~\ref{fig:architecture}), which is set identical to the current generator (i.e, S2S model), to conduct the sampling conditioned on $y_t$ (that is, perform generation with the starting token as $y_t$). Let $\mathrm{\bf \hat{Y}}_t$ be the sampled response, then, we employ the pre-trained rank-aware calibrator to calculate the fine-grained quality score for $\mathrm{\bf \hat{Y}}_t$. Specifically, $\mathrm{\bf \hat{Y}}_t$ is fed to the components $\mathrm{D}^{r}$, $\mathrm{D}^{pt}$ and $\mathrm{D}^{pair}$ separately and the final quality score is the mean value of the outputs from the scoring models:
\begin{equation}
\small
\label{eq:token}
    \mathrm{q}_t = \frac{1}{3}[\mathrm{D}^{r}(\mathrm{\bf X}, \mathrm{\bf \hat{Y}}_t) + \mathrm{D}^{pt}(\mathrm{\bf X}, \mathrm{\bf \hat{Y}}_t) + \mathrm{D}^{pair}(\mathrm{\bf X}, \mathrm{\bf \hat{Y}}_t)]
\end{equation}
To reduce the variance of the sampling, we perform the Monte-Carlo search $N$ times, yielding a set of simulated responses $\{\mathrm{\bf \hat{Y}}^{1}_t,\cdots,\mathrm{\bf \hat{Y}}^{N}_t\}$ for the response token $y_t$. The final token-level quality estimations for $y_t$ is:
\begin{equation}
\setlength{\abovedisplayskip}{3pt}
\setlength{\belowdisplayskip}{3pt}
    \mathrm{q}_t = \frac{1}{N}\sum^{N}_{i=1} \mathrm{q}^{i}_t  
\end{equation}
where $\mathrm{q}^{i}_t$ is the fine-grained quality score of the sampled response $\mathrm{\bf \hat{Y}}^{i}_t$. Here, $\mathrm{q}_t$ characterizes the contribution of the $t$-th generation step to computing the training sample loss. Compared to~\cite{shang2018learning,liu-etal-2018-towards-less}, which treat each generation step equally, performing the weighted sum of the losses at each generation step can alleviate the incorrect bonus on some uninformative words. The calculations of the sample loss $\mathcal{L}_{\text{RC}}(\mathrm{\bf X}, \mathrm{\bf Y})$ and the total loss $\mathcal{L}_{\text{RC}}$ are given below:
\begin{equation*}
\setlength{\abovedisplayskip}{3pt}
\setlength{\belowdisplayskip}{3pt}
\small
\begin{split}
\mathcal{L}_{\text{RC}}(\mathrm{\bf X}, \mathrm{\bf Y}) &= -\sum^{m}_{t=1} \mathrm{q}_t * \log(p(y_t|\mathrm{\bf X},y_{1:t-1})) \\ 
\mathcal{L}_{\text{RC}} &= \frac{1}{|\mathbb{D}|} \sum_{(\mathrm{\bf X}, \mathrm{\bf Y}) \in \mathbb{D}} \mathcal{L}_{\text{RC}}(\mathrm{\bf X}, \mathrm{\bf Y})
\end{split}
\end{equation*}

\subsection{Knowledge Inference}
Recall the issue that some meaningful words in the irrelevant responses may be erroneously penalized due to the topical irrelevance. For example, in the fifth response in Table~\ref{tab:example}, the tokens ``university'' and ``martyrs'' will be assigned low quality scores and thus the generation of these words will be suppressed after the training. To reduce such kind of side effect, we propose Knowledge Inference (KI) component. The key idea of this component is to guide the generation with the useful knowledge. In this paper, we regard \textbf{keywords} (i.e., informative words) as the knowledge. Firstly, we utilize TextRank model~\cite{mihalcea-tarau-2004-textrank} to extract the keywords from the responses in the training dataset. Then, we construct a knowledge memory $\mathrm{\bf K} \in \mathbb{R}^{n_k \times \mathrm{dim}_k}$, where $n_k$ memory vectors correspond to $n_k$ informative words obtained from the training corpus. After building the knowledge memory, we leverage the context-aware decoding hidden representation $\mathrm{\bf s}_t$ to pay attention to the memory vectors $\mathrm{\bf k}_j$ ($j \in (1, \ldots, n_k)$) and summarize the knowledge via weighted combination:
\begin{equation}
\setlength{\abovedisplayskip}{3pt}
\setlength{\belowdisplayskip}{3pt}
\begin{split}
    p(\mathrm{\bf k}_j|\mathrm{\bf s}_t) &= \text{tanh}(\mathrm{\bf s}_t ^\top \mathrm{\bf W}_K \mathrm{\bf k}_j) \\
    \mathrm{\bf c}^{k}_t &= \frac{1}{n_k} \sum^{n_k}_{j=1} p(\mathrm{\bf k}_j|\mathrm{\bf s}_t) \mathrm{\bf k}_j
\end{split}
\end{equation}
where $\mathrm{\bf W}_K$ is a parameter matrix in the KI component and $\text{tanh}$ denotes the hyperbolic tangent function. $\mathrm{\bf c}^{k}_t$ is the summarized knowledge at the $t$-th generation timestep. To incorporate the knowledge into the generation, we feed the concatenated $\mathrm{\bf c}^{k}_t$ and decoder hidden representation $\mathrm{\bf s}_t$ to the ultimate softmax layer for producing the token $y^{'}_t$. 

Learning the alignment between the memory $\mathrm{\bf K}$ and the decoder hidden representation $\mathrm{\bf s}_t$ from scratch is difficult because the model has no prior knowledge of the informative words. Thus, we regard the extracted informative words from the current ground-truth response $\mathrm{\bf Y}$ as additional supervision signal to help the training of $\mathrm{\bf K}$. Given the informative words $\mathrm{\bf Y}^{K}$ extracted from $\mathrm{\bf Y}$, the objective of the KI component is defined as follows:
\begin{equation}
\setlength{\abovedisplayskip}{3pt}
\setlength{\belowdisplayskip}{3pt}
\small
    \mathcal{L}_{\text{KI}} = -\frac{1}{|\mathbb{D}|} \sum_{(\mathrm{\bf X}, \mathrm{\bf Y}) \in \mathbb{D}} \sum^{|\mathrm{\bf Y}|}_{t=1} \sum^{n_k}_{j=1} \mathrm{b}^{K}_j * \delta(p(\mathrm{\bf k}_j|\mathrm{\bf s}_t))
\end{equation}
\[
    \mathrm{b}^{K}_j = 
    \begin{cases} 
      1, & j \in \mathrm{\bf Y}^{K} \\
      0, & \text{otherwise}
   \end{cases}
\]
where the binary variable $\mathrm{b}^{K}_j$ is the indicator of the existence of the $j$-th informative word and $\delta$ denotes the \texttt{sigmoid} activation function. $p(\mathrm{\bf k}_j|\mathrm{\bf s}_t)$ is the alignment score between $\mathrm{\bf k}_j$ and $\mathrm{\bf s}_t$. By introducing the objective $\mathcal{L}_{\text{KI}}$, the model can learn better knowledge memory $\mathrm{\bf K}$ and its alignment with $\mathrm{\bf s}_t$ based on more feedback from both the S2S component and the KI component. Consequently, there will be a higher potential generating informative words.

\subsection{Joint Training}
The parameters of the pre-trained rank-aware calibrator model are kept fixed all the time. Although the loss with rank-aware calibration, i.e., $\mathcal{L}_{\text{RC}}$, reduce the possibility of generating unimportant words, it may weaken the language model constraint on the generated responses. Thus, we combine the NLL 
loss $\mathcal{L}_{\text{NLL}}$ with $\mathcal{L}_{{RC}}$ during the training. The loss $\mathcal{L}_{\text{KI}}$ is also included to encourage the generation of the informative words. The final training objective $\mathcal{J}(\theta)$ of the proposed framework is as follows:
\begin{equation}
\setlength{\abovedisplayskip}{3pt}
\setlength{\belowdisplayskip}{3pt}
    \mathcal{J}(\theta) = \mathcal{L}_{\text{NLL}} + \mathcal{L}_{\text{RC}} + \mathcal{L}_{\text{KI}}
\end{equation}

\begin{table*}[]
\renewcommand{\arraystretch}{0.9}
    \centering
    \resizebox{1.8\columnwidth}{!}{
    \begin{tabular}{ll|ccc|ccc|cc}
    \Xhline{3\arrayrulewidth}
     & \multirow{2}{*}{\textbf{Model}} & \multicolumn{3}{c|}{\textbf{Word Overlap}} & \multicolumn{3}{c|}{\textbf{Embedding Similarity}} & \multicolumn{2}{c}{\textbf{Diversity}} \\ \cline{3-10}
    & & BLEU-1 & BLEU-2 & BLEU-3 & Average & Extrema & Greedy &  Dist-1 & Dist-2 \\ \hline \hline
    \multirow{7}{*}{\textbf{w/o MMI}} & S2S & 17.3 & 3.9 & 1.3 & 0.481 & 0.286 & 0.403 & 0.059 & 0.232 \\ 
    & S2S-RW & 17.1 & 3.4 & 1.1 & 0.519 & 0.294 & 0.409 & 0.052 & 0.213 \\
    & S2S-CN & 18.1 & 4.0 & 1.0 & 0.525 & 0.302 & 0.401 & 0.058 & 0.224  \\
    & S2S-GT & 18.9 & 4.3 & 1.5 & 0.521 & 0.317 & 0.411 & 0.062 & 0.245 \\ 
    & S2S-DF & 16.9 & 3.0 & 0.8 & 0.491 & 0.295 & 0.398 & 0.060 & 0.238\\
    & $\text{S2S-DF}^{+}$ & 20.2 & 4.5 & 1.7 & 0.501 & 0.308 & 0.417 & 0.054 & 0.213 \\
    & OURS & \textbf{20.9} & \textbf{4.9} & \textbf{1.8} & \textbf{0.559} & \textbf{0.342} & \textbf{0.471} & \textbf{0.077} & \textbf{0.288} \\ \hline \hline
    \multirow{6}{*}{\textbf{w/ MMI}} & S2S & \textbf{22.0} & 4.8 & 1.6 & 0.530 & 0.308 & 0.395 & 0.076 & 0.321 \\
    & S2S-RW & 20.2 & 4.3 & 1.4 & 0.519 & 0.318 & 0.400 & 0.072 & 0.309 \\ 
    & S2S-CN & 21.0 & 4.4 & 1.4 & 0.517 & 0.313 & 0.398 & 0.066 & 0.310 \\
    & S2S-GT & 21.5 & 4.7 & 1.6 & 0.523 & 0.306 & 0.410 & 0.076 & 0.318 \\ 
    & S2S-DF & 21.0 & 4.5 & 1.5 & 0.510 & 0.299 & 0.408 & 0.070 & 0.315 \\
    & $\text{S2S-DF}^{+}$ & 21.4 & 4.6 & 1.5 & 0.520 & 0.304 & 0.415 &0.072 & 0.307 \\ 
    & OURS & \textbf{22.0} & \textbf{4.8} & \textbf{1.7}  & \textbf{0.563} & \textbf{0.347} & \textbf{0.478} & \textbf{0.095} & \textbf{0.377}  \\
    \Xhline{3\arrayrulewidth}
    \end{tabular}}
    \caption{Results on automatic metrics.}
    \label{tab:automatic}
    \vspace{-3mm}
\end{table*}

\section{Experimental Setup}
\subsection{Dataset}
To investigate the effectiveness of the proposed framework, we build a dialogue dataset with multi-level fine-grained quality labels. As shown in Table~\ref{tab:example}, we consider 5-level response quality: (1) L5: relevant and interesting; (2) L4: relevant but simple; (3) L3: acceptable but universal; (4) L2: quiet boring; (5) L1: irrelevant. To process such quality labels, we convert L1 to L5 to the normalized quality scores, namely, 0.0, 0.25, 0.5, 0.75 and 1.0 respectively.

We crawl 534,381 query-response pairs from \textit{Douban Group}, a famous forum in China\footnote{https://www.douban.com/group/explore}, and recruit human annotators to annotate the multi-level quality labels for each sample. We split the annotated dataset into three parts, where 522,881 query-response pairs are for training, 10000 query-response pairs are for validation and the remaining 1500 query-response pairs are for testing. There is no overlap among the queries in training, validation and testing set.

\subsection{Settings}
We compare our model with the following baselines and comparison models:

\noindent $\bullet$   \textbf{S2S}: It is the standard attention-based sequence-to-sequence model~\cite{bahdanau2015neural}. \\
\noindent $\bullet$   \textbf{S2S-RW}~\cite{liu-etal-2018-towards-less}: \textbf{S2S} with \textbf{R}e-\textbf{W}eighting. It is an extended S2S model with response quality measurement based on the corpus-level n-gram co-occurrence statistics and the length of the response. \\
\noindent $\bullet$   \textbf{S2S-CN}~\cite{shang2018learning}: \textbf{S2S} with \textbf{C}alibration \textbf{N}etworks. It is an enhanced S2S model where a pre-trained calibration networks is introduced to adjust the weight of the sample loss. \\
\noindent $\bullet$   \textbf{S2S-GT}: \textbf{S2S} with \textbf{G}round \textbf{T}ruth. It is a variant of the models~\cite{liu-etal-2018-towards-less} and~\cite{shang2018learning}, where we replace the instance quality measurement with the ground truth quality scores. \\
\noindent $\bullet$   \textbf{S2S-DF}: \textbf{S2S} with \textbf{D}ata \textbf{F}iltering. It is a S2S model trained on the high quality data where we adopt the data filtering strategy proposed in~\cite{xu-etal-2018-better} to obtain the topically related query-response pairs.\footnote{ https://github.com/XinnuoXu/CVAE\_Dial.} \\
\noindent $\bullet$   $\textbf{S2S-DF}^{+}$: A variant of \textbf{S2S-DF}. We perform data filtering according to the gold standard multi-level quality score. Specifically, we only preserve query-response pairs of L4 or L5 quality for training the model.\footnote{By doing this, the size of the training dataset reduces from 522,881 to 334,471.}

We divide the comparisons with the baseline models into two groups according to using the Maximum Mutual Information (\textbf{MMI}) decoding~\cite{li-etal-2016-diversity} or not. The first group of comparison is under the setting \textbf{w/o MMI}, where the decoding is identical to the normal beam search. The second group is under the setting \textbf{w/ MMI}, where an inverse S2S (response-to-query) model is introduced to rerank the $N$-best ($N$ is set as 50) hypothesis generated from the standard S2S (query-to-response) model.
We employ a two-layer bi-directional LSTM as the encoder and a two-layer unidirectional LSTM as the decoder.
The dimension $\mathrm{dim}_h$ of both the encoder and the decoder hidden representations is 500. The word embeddings are randomly initialized and the size of each word vector is 300. For the model training, we employ Adam~\cite{kingma2014adam} as optimizer, with the initial learning rate being 0.0001 and the decay rate being 0.9. The dimension of the keyword memory vector $\mathrm{dim}_k$ is 30.
In the rank-aware calibrator, we employ multi-channel CNN~\cite{kim-2014-convolutional} to extract sentence-level features. The sizes of convolutional filter are 1, 2, 3 and the number of filters is 100 for each size.

\begin{table}[!t]
\renewcommand{\arraystretch}{0.95}
    \centering
    \resizebox{0.8\columnwidth}{!}{
    \begin{tabular}{l|cccc}
    \Xhline{2.5\arrayrulewidth}
        Model & \textbf{+2} & \textbf{+1} & \textbf{0} & Avg.  \\ \hline
        S2S & 7.50 & 61.00 & 31.50 & 76.00 \\ 
        S2S \textbf{w/ MMI} & 10.00 & 59.50 & 30.50 & 79.50 \\ 
        S2S-GT \textbf{w/ MMI} & 10.50 & 60.00 & 29.50 & 81.00 \\ 
        OURS \textbf{w/ MMI} & 13.50 & 58.00 & 28.50 & 85.00 \\ \Xhline{2.5\arrayrulewidth}
    \end{tabular}}
    \caption{Human evaluation results. (\%)}
    \label{tab:human}
    \vspace{-3mm}
\end{table}

\subsection{Evaluation Metrics}
We evaluate our model and the comparison models using the following evaluation metrics:\\
\noindent{\textbf{Word Overlap:}} we employ BLEU-1, BLEU-2 and BLEU-3 to measure the word overlap between the generated and the gold standard responses.\\
\noindent{\textbf{Embedding Similarity:}} Actually, BLEU metric does not correlate strongly with human judgements. Thus, we introduce the embedding-based metrics~\cite{liu-etal-2016-evaluate} to measure the similarity between the generated results and the ground truth. \\
\noindent{\textbf{Diversity:}} Following~\cite{li-etal-2016-diversity}, we calculate the ratios of the distinct unigrams (Dist-1) and bigrams (Dist-2) in the generated responses, and use the metrics to measure how diverse and informative the responses are.\\
\noindent{\textbf{Human Evaluation:}} Apart from the automatic evaluations, we randomly sample 200 queries in the testing dataset and recruit three helpers to judge the quality of each generated response from the best models on the automatic metrics. The rating criteria, which is same as that in~\cite{shang2018learning}, is as follows: \textbf{+2}: The response is not only relevant and natural, but also informative and interesting. \textbf{+1}: The response can be used as a reply to the message, but is too universal like ``Yes, I see'' , ``Me too'' and ``I don’t know''. \textbf{0}: The response cannot be used as a reply to the message. It is either semantically irrelevant or disfluent. For each model, we report the ratio of each score (\textbf{+2}, \textbf{+1} or \textbf{0}) and the average score as the human evaluation results. 
\begin{table}[!t]
    \centering
    \resizebox{1\columnwidth}{!}{
    \begin{tabular}{l|ccc|cc}
    \Xhline{3\arrayrulewidth}
    \multirow{2}{*}{Model} & \multicolumn{3}{c|}{\textbf{Embedding Similarity}} & \multicolumn{2}{c}{\textbf{Diversity}} \\ \cline{2-6}
    & Average & Extrema & Greedy &  Dist-1 & Dist-2 \\ \hline
    OURS & \textbf{0.559} & \textbf{0.342} & \textbf{0.471} & \textbf{0.077} & \textbf{0.288}  \\
    OURS w/o KI & 0.521 & 0.311 & 0.435 & 0.068 & 0.261 \\
    OURS w/o RC & 0.497 & 0.275 & 0.418 & 0.066 & 0.267 \\
    OURS w/o RC \& KI & 0.481 & 0.286 & 0.403 & 0.059 & 0.232 \\
    \Xhline{3\arrayrulewidth} 
    \end{tabular}}
    \caption{Results of ablation study.}
    \label{tab:ablation}
    \vspace{-3mm}
\end{table}

\begin{table*}[!t]
    \centering
    \resizebox{2.1\columnwidth}{!}{
    \begin{tabular}{l|l|l|l}
    \Xhline{3\arrayrulewidth}
        Query & S2S \textbf{w/ MMI} & S2S-GT \textbf{w/ MMI} & OURS \textbf{w/ MMI} \\ \hline
        \begin{CJK}{UTF8}{gbsn}1. 夜深人静了\end{CJK} & \begin{CJK}{UTF8}{gbsn}嗯, 是的\end{CJK} & \begin{CJK}{UTF8}{gbsn}我也想吃\end{CJK} & \begin{CJK}{UTF8}{gbsn}我还没\textbf{\textcolor{blue}{[}\underline{\textcolor{blue}{睡觉}}\textcolor{blue}{]}}\end{CJK} \\
        All is quiet at dead of night. & Oh, yes. & I want to eat too. & But I haven't gone to \textbf{\textcolor{blue}{[}\underline{\textcolor{blue}{sleep}}\textcolor{blue}{]}} yet. \\ \hline
        \begin{CJK}{UTF8}{gbsn}2. 说说处处们的初恋-最爱-被爱都是什么星座的\end{CJK} & \begin{CJK}{UTF8}{gbsn}我是\textbf{\textcolor{blue}{[}\underline{\textcolor{blue}{摩羯}}\textcolor{blue}{]}}\end{CJK} & \begin{CJK}{UTF8}{gbsn}我也是\textbf{\textcolor{blue}{[}\underline{\textcolor{blue}{天蝎}}\textcolor{blue}{]}}\end{CJK} & \begin{CJK}{UTF8}{gbsn}\textbf{\textcolor{blue}{[}\underline{\textcolor{blue}{初恋}}\textcolor{blue}{]}}是\textbf{\textcolor{orange}{[}\underline{\textcolor{orange}{天秤座}}\textcolor{red}{]}}\end{CJK} \\ 
        Tell me the constellations of your first love, your best love and your suitor & I am \textbf{\textcolor{blue}{[}\underline{\textcolor{blue}{Capricorn}}\textcolor{blue}{]}}. & My constellation is also \textbf{\textcolor{blue}{[}\underline{\textcolor{blue}{Scorpio}}\textcolor{blue}{]}}. & The constellation of my \textbf{\textcolor{blue}{[}\underline{\textcolor{blue}{first love}}\textcolor{blue}{]}} is \textbf{\textcolor{orange}{[}\underline{\textcolor{orange}{Libra}}\textcolor{orange}{]}}. \\ \hline
        \begin{CJK}{UTF8}{gbsn}3. 你玩的第一个网游是什么?别吹\end{CJK} & \begin{CJK}{UTF8}{gbsn}我玩的\textbf{\textcolor{blue}{[}\underline{\textcolor{blue}{游戏}}\textcolor{blue}{]}}\end{CJK} & \begin{CJK}{UTF8}{gbsn}最近玩的\textbf{\textcolor{blue}{[}\underline{\textcolor{blue}{游戏}}\textcolor{blue}{]}}\end{CJK} & \begin{CJK}{UTF8}{gbsn}\textbf{\textcolor{blue}{[}\underline{\textcolor{blue}{梦幻}}\textcolor{blue}{]}}\textbf{\textcolor{orange}{[}\underline{\textcolor{orange}{网游}}\textcolor{orange}{]}}!!\end{CJK} \\ 
        What is the first online game you have ever played? & The \textbf{\textcolor{blue}{[}\underline{\textcolor{blue}{game}}\textcolor{blue}{]}} I am playing. & The \textbf{\textcolor{blue}{[}\underline{\textcolor{blue}{game}}\textcolor{blue}{]}} I recently played. & An \textbf{\textcolor{orange}{[}\underline{\textcolor{orange}{online game}}\textcolor{orange}{]}} called \textbf{\textcolor{blue}{[}\underline{\textcolor{blue}{Fantasy Westward Journey}}\textcolor{blue}{]}} !! \\ \hline
        \begin{CJK}{UTF8}{gbsn}4. 我曾经跨过山和大海\end{CJK} & \begin{CJK}{UTF8}{gbsn}我也想知道\end{CJK} & \begin{CJK}{UTF8}{gbsn}我也想和他在一起\end{CJK} & \begin{CJK}{UTF8}{gbsn}我一直很喜欢这句\textbf{\textcolor{blue}{[}\underline{\textcolor{blue}{话}}\textcolor{blue}{]}}\end{CJK} \\ 
        I've crossed the mountains and the sea. & I also want to know & I want to be with him too. & It is the \textbf{\textcolor{blue}{[}\underline{\textcolor{blue}{quotation}}\textcolor{blue}{]}} I've always loved \\ \hline
        \begin{CJK}{UTF8}{gbsn}5. 你用的什么浏览器\end{CJK} & \begin{CJK}{UTF8}{gbsn}我用的是\textbf{\textcolor{blue}{[}\underline{\textcolor{blue}{浏览器}}\textcolor{blue}{]}}\end{CJK} & \begin{CJK}{UTF8}{gbsn}我用的是\textbf{\textcolor{blue}{[}\underline{\textcolor{blue}{浏览器}}\textcolor{blue}{]}}\end{CJK} & \begin{CJK}{UTF8}{gbsn}用的是\textbf{\textcolor{blue}{[}\underline{\textcolor{blue}{百度}}\textcolor{blue}{]}}\textbf{\textcolor{orange}{[}\underline{\textcolor{orange}{浏览器}}\textcolor{orange}{]}}\end{CJK} \\ 
        Which browser do you use. & What I use is \textbf{\textcolor{blue}{[}\underline{\textcolor{blue}{browser}}\textcolor{blue}{]}} & What I use is browser. & The used \textbf{\textcolor{orange}{[}\underline{\textcolor{blue}{browser}}\textcolor{orange}{]}} is \textbf{\textcolor{blue}{[}\underline{\textcolor{blue}{Baidu}}\textcolor{blue}{]}}. \\ \hline
    \Xhline{3\arrayrulewidth} 
    \end{tabular}}
    \caption{Example output, color printing is preferred. The informative words are wrapped in brackets and coupled with underline.}
    \label{tab:case_analysis}
    \vspace{-3mm}
\end{table*}

\section{Results and Discussions}
\subsection{Main Results}
Table~\ref{tab:automatic} depicts the experimental results on the automatic metrics. Under the settings of both \textbf{w/o MMI} (the first group in Table~\ref{tab:automatic}) and \textbf{w/ MMI} (the second group in Table~\ref{tab:automatic}), our model consistently outperforms the baseline and the comparison models on all of the metrics, demonstrating the effectiveness of the proposed components. Specifically, our model is better than all of the non MMI-based baselines on Dist-1 and Dist-2, suggesting that the proposed model can generate more diverse and informative responses. The probable reason is that we introduce a knowledge inference component to encourage the generation of some informative words erroneously punished by the models adopting instance-weighting strategies (i.e., S2S-RW, S2S-CN and S2S-GT). We also observe that OURS achieves better performance on BLEU scores and Embedding Similarity compared with S2S-GT. This finding shows that token-level training loss estimation together with rank-aware calibration is a better strategy than just scaling the sample loss with an instance weight. Another interesting finding is that training model with only the high-quality data, as done in S2S-DF and $\text{S2S-DF}^{+}$, is still worse than the instance weighting strategy adopted in S2S-GT, indicating that soft data filtering via quality estimation is a more suitable way for integrating quality information into the model training. 

From the human evaluation results in Table~\ref{tab:human}, our model generates the most satisfactory responses (responses labeled as \textbf{+2}) and the least invalid responses (responses labeled as \textbf{0}). Specifically, OURS \textbf{w/ MMI} increases 3.5\% \textbf{+2} responses and reduces 2.0\% \textbf{0} responses compared with S2S \textbf{w/ MMI}. Without introducing the additional knowledge and diversity modeling, the pure S2S model tend to produce more generic responses than the MMI-based models, i.e., S2S \textbf{w/ MMI} and OURS \textbf{w/ MMI}. We also find that taking the instance quality into consideration can lead to less generation of the \textbf{0} responses, for example, OURS \textbf{w/ MMI} / S2S-GT \textbf{w/ MMI} versus S2S \textbf{w/ MMI}. The above observations are somewhat consistent with those in the Table~\ref{tab:automatic}. 

\subsection{Ablation Study}
To further investigate the effectiveness of the proposed component, we compare our model with the ablated models. The results of the ablation study are listed in Table~\ref{tab:ablation}. OURS significantly outperforms the ablated models on the relevance metrics (i.e., Embedding Similarity) but the performance gap on the diversity metrics is small. It is reasonable because the aim of the proposed components is not to model the diversity. Removing either the KI component or the RC component will result in performance degradation, especially for the case removing the RC component, our model degenerates to the S2S model together with the knowledge inference (KI) component. Comparing with OURS, the relevance scores of the OURS w/o RC drastically decrease (-6\%, -6\%, -7\% on Average, Extrema and Greedy respectively), indicating that our token-level rank-aware calibration is helpful for improving the relevance of the generated responses. Without RC \& KI components, our model will become equivalent to the standard S2S model and the metric scores will drop for further step.

\begin{figure}[!t]
    \centering
    \includegraphics[width=1\columnwidth]{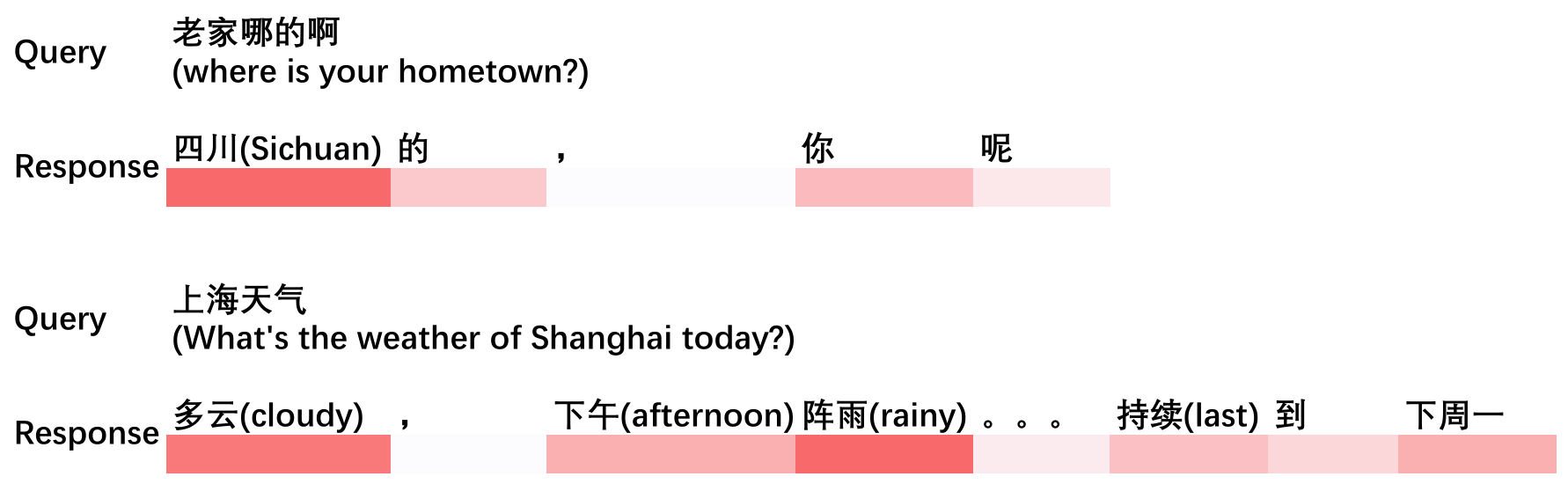}
    \caption{Example output for token-level quality estimation (i.e., $q_t$ in Equation~\ref{eq:token}), ``Response'' refers to the gold standard response. Better viewed in color.}
    \label{fig:rc}
     \vspace{-3mm}
\end{figure}

\subsection{Case Visualization and Analysis}
In Figure~\ref{fig:rc}, we visualize the token-level quality scores calculated from Eq~\ref{eq:token}. Each cell corresponds to a Chinese word and it uses depth of color to represent the importance of this word. We can observe that the importance score for the token ``Sichuan'', a province in China, is significantly larger than those for other tokens when the query is talking about ``hometown'' (the upper query). Similarly, the token ``cloudy'' and ``rainy'' are highlighted by the rank-aware calibrator when the topic of the query is related to the weather. These examples demonstrate that our token-level strategy armed with Monte-Carlo Rollout can accurately characterize the quality of token with respect to the query, leading to better calibration for the training. 

Table~\ref{tab:case_analysis} shows some example outputs. The informative words appearing in the training dataset are wrapped in the brackets and coupled with underlines. According to these cases, we find that our model (i.e., OURS \textbf{w/ MMI}) can produce some interesting responses with the meaningful words. For example, the response for the third query mentions ``Fantasy Westward Journey'', the name of a very popular online game in China. Similarly, the response answers ``Baidu'', a Chinese IT company, for the fifth query. We attribute these phenomena to the KI component, which explicitly encourages the generation of the important words. Besides, our model can give replies closely relevant to some queries without explicit topic, for example, the first and the the fourth query, it is probably because our model introduces fine-grained quality information via token-level rank-aware calibration and the probability of generating generic response such as ``oh, yes'' and ``I also want to know'' is thus reduced.

\section{Related Work}
Neural Dialogue Generation is usually formulated as a sequence translation problem~\cite{ritter-etal-2011-data,shang-etal-2015-neural,serban2016building,wang2018chat} and the sequence-to-sequence (S2S) encoder-decoder framework~\cite{cho-etal-2014-learning,sutskever2014sequence,bahdanau2015neural} is applied. Various approaches have been proposed to improve the S2S model for better human-computer conversation, for example, via introducing topic or keyword information into the generation process~\cite{mou-etal-2016-sequence,xing2017topic,gao2019generating}, diversifying the generated responses with additional memories or objectives~\cite{li-etal-2016-diversity,zhou2018elastic}, and arming the S2S model with GAN or other advanced techniques~\cite{xu-etal-2017-neural,du-etal-2018-variational,tao2018get,cai-etal-2019-skeleton,gao-etal-2019-discrete}.

Different from the above works, \cite{shang2018learning,liu-etal-2018-towards-less,xu-etal-2018-better,csaky-etal-2019-improving,cai-etal-2020-data} attempt to incorporate the quality modeling of training data into S2S training. The basic idea of these approaches is either instance weighting~\cite{jiang-zhai-2007-instance} or filtering data of low quality~\cite{wojciechowski2002dataset}. The potential issue of these strategies is that the tokens within each training instance are treated equally and the informative words in a low-quality response may be erroneously punished. Similarly, the model guided by these strategies tends to generate some boring but frequently-used words.  


\section{Conclusion}
We propose a multi-level contrastive learning paradigm to exploit the fine-grained response quality to calibrate the training of the response generation models. We design a Rank-aware Calibration (RC) network to construct the contrastive optimization objectives. We further build a Knowledge Inference (KI) component to capture the keyword knowledge from the reference during training and exploit such information to encourage the generation of informative words. We evaluate the proposed model on a carefully annotated short-text conversation dataset and the results suggest that our model can generate more relevant and diverse responses compared to the baseline models.

\begin{small}
\bibliographystyle{named}
\bibliography{ijcai21}
\end{small}

\end{document}